\documentclass[10pt,twocolumn,letterpaper]{article}

\usepackage{cvpr}              %

\usepackage{graphicx}
\usepackage{amsmath}
\usepackage{amssymb}
\usepackage{booktabs}

\usepackage{dsfont}
\usepackage{bm}
\usepackage{multirow}
\usepackage{xcolor}
\makeatletter
\@namedef{ver@everyshi.sty}{}
\makeatother
\usepackage{tikz, graphicx}
\let\svtikzpicture\tikzpicture
\def\tikzpicture{\noindent\svtikzpicture}

\usepackage{standalone}
\usepackage{svg}
\usepackage{adjustbox}

\usepackage{comment}
\usepackage{array}
\usepackage{enumitem}%

\makeatletter
\newcommand\notsotiny{\@setfontsize\notsotiny\@vipt\@viipt}
\makeatother

\usepackage[thinlines]{easytable}

\usepackage[T1]{fontenc}

\newcommand\Tstrutsmall{\rule{0pt}{2ex}} 
\newcommand\Tstrut{\rule{0pt}{11.5ex}} 
\newcommand\Tstrutt{\rule{0pt}{12.7ex}} 
\newcommand\Tstrutsuper{\rule{0pt}{35.7ex}}
\newcommand\Tstrutarch{\rule{0pt}{33ex}}

\newlength{\Oldarrayrulewidth}
\newcommand{\Cline}[2]{%
  \noalign{\global\setlength{\Oldarrayrulewidth}{\arrayrulewidth}}%
  \noalign{\global\setlength{\arrayrulewidth}{#1}}\cline{#2}%
  \noalign{\global\setlength{\arrayrulewidth}{\Oldarrayrulewidth}}}

\newcommand{\imgres}[1]{#1_lr} %

\usepackage[pagebackref=true,breaklinks=true,letterpaper=true,colorlinks=true,bookmarks=false]{hyperref}

\usepackage[capitalize]{cleveref}
\crefname{section}{Sec.}{Secs.}
\Crefname{section}{Section}{Sections}
\Crefname{table}{Table}{Tables}
\crefname{table}{Tab.}{Tabs.}

\def\cvprPaperID{12} %
\def\confName{EarthVision}
\def\confYear{2022}

\begin{document}

\title{Sat-NeRF: Learning Multi-View Satellite Photogrammetry \\ With Transient Objects and Shadow Modeling Using RPC Cameras}

\author{Roger Mar\'i \quad Gabriele Facciolo \quad Thibaud Ehret\\
Université Paris-Saclay, CNRS, ENS Paris-Saclay, Centre Borelli, 91190, Gif-sur-Yvette, France\\
{\href{https://centreborelli.github.io/satnerf}{\tt\small https://centreborelli.github.io/satnerf}}
}
\maketitle

\begin{abstract}
   We introduce the Satellite Neural Radiance Field (Sat-NeRF), a new end-to-end model for learning multi-view satellite photogrammetry in the wild. Sat-NeRF combines some of the latest trends in neural rendering with native satellite camera models, represented by rational polynomial coefficient (RPC) functions. The proposed method renders new views and infers surface models of similar quality to those obtained with traditional state-of-the-art stereo pipelines. Multi-date images exhibit significant changes in appearance, mainly due to varying shadows and transient objects (cars, vegetation). Robustness to these challenges is achieved by a shadow-aware irradiance model and uncertainty weighting to deal with transient phenomena that cannot be explained by the position of the sun. We evaluate Sat-NeRF using WorldView-3 images from different locations and stress the advantages of applying a bundle adjustment to the satellite camera models prior to training. This boosts the network performance and can optionally be used to extract additional cues for depth supervision.
\end{abstract}

\section{Introduction}
\label{sec:intro}

High-resolution satellite imagery is a valuable resource for countless economic activities, many of them based on knowledge of the geometry of the Earth's surface and its changes. This has triggered the development of a number of pipelines capable of highly accurate depth estimation from disparity using multiple satellite views \cite{dangelo2012dense, defranchis2014automatic, facciolo2017automatic, rupnik20183d, gong2019dsm, shean2016automated, beyer2018ames}. The output large-scale 3D models are usually represented using discrete point clouds or digital surface models (DSMs) of a certain resolution.

\input{fig_teaser}

The latest works in 3D modeling from multiple views show that it is possible to achieve a superior representation of a 3D object or scene by \textit{learning} it as a continuous function or field $\mathcal{F}$ \cite{tewari2020state}. Neural rendering methods learn $\mathcal{F}$ by integrating differentiable rendering techniques into a neural network. The tasks of novel view synthesis and 3D reconstruction are then solved implicitly, as the network is trained to figure out which geometry and color radiances fit the camera projection mappings of the different views. 

Neural radiance fields (NeRFs) have gained great popularity in the field of neural rendering \cite{mildenhall2020nerf, tewari2020state}. In this paper, we introduce a NeRF variant architecture that achieves state-of-the-art results in novel view synthesis and 3D reconstruction from high-resolution satellite imagery in the wild. We refer to our variant as Satellite NeRF or Sat-NeRF. The original NeRF approach is not adapted to satellite images, e.g. because of the specificities of the camera models, the large distance between the cameras and the scene or the appearance inconsistencies of multi-date collections \cite{martin2021nerf, derksen2021shadow}. Sat-NeRF addresses these challenges using some of the latest advances in NeRFs \cite{martin2021nerf, derksen2021shadow, deng2021depth} and adapting well-known
tools for satellite image processing. As a result, the model learns highly accurate 3D geometry, similar to that obtained with state-of-the-art stereo pipelines relying on handcrafted features (Figure~\ref{fig:teaser}).

Our contributions consist of:
\begin{itemize}\setlength\itemsep{-0.2em}
    \item[-] A NeRF variant that combines existing state-of-the-art methods to adapt to the satellite context. It is robust to the radiometric inconsistencies of multi-date satellite imagery, comprising shadows caused by a single non-static light source (the sun) and small transient objects (mainly trees or cars in open-air parkings).
    \item[-] A point sampling strategy adapted to satellite camera models. The rational polynomial camera (RPC) model \cite{grodecki2001ikonos, fraser2006sensor} of each input image is directly used to cast rays in the object space. This RPC-based strategy provides independence to the satellite system and improves the results obtained with approximate pinhole cameras.
    \item[-] A study of the advantages of correcting RPC inconsistencies before training, e.g. by means of a bundle adjustment \cite{mari2021generic}. We show that eluding this step leads to a drop in the performance of the model. In addition, we detail how to reuse the sparse point cloud employed in the bundle adjustment to improve geometry learning.
\end{itemize}

We evaluate Sat-NeRF on different areas of interest covering $256 \! \times \! 256$ m each, using $\sim$10-20 RGB crops from multi-date WorldView-3 images for training \cite{bosch2019semantic, le20192019}. A lidar digital surface model (DSM) of resolution 0.5 m/pixel is used as ground truth model to assess the geometry. Sat-NeRF is compared to other NeRF variants~\cite{mildenhall2020nerf, derksen2021shadow} as well as a state-of-the-art traditional satellite stereo pipeline~\cite{defranchis2014automatic}. We also publish the code and data used for this article.
\section{Related work}
\label{sec:related_work}

Current state-of-the-art 3D reconstruction pipelines for satellite images typically follow multi-view stereo approaches, which can outperform sophisticated true multi-view software \cite{ozcanli2015comparison, gomez2022experimental}. Due to the complexity of the task, satellite stereo pipelines can still be improved in a number of aspects. Some of the most important limitations are:
\begin{itemize}\setlength\itemsep{-0.1em}
    \item[-] The 3D reconstruction usually follows the estimation of a dense disparity map using matching strategies derived from the semi-global matching algorithm \cite{hirschmuller2008stereo, dangelo2012dense, defranchis2014automatic, gong2019dsm, beyer2018ames}. Therefore, human-crafted features and cost functions are at the core of the methodology.
    \item[-] The selection of suitable stereo pairs to estimate disparity is another major challenge. Criteria based on image metadata (e.g. acquisition dates, incidence angles, etc.) have proven to be useful, but do not guarantee the best choice \cite{facciolo2017automatic, gong2019dsm, han2020state}.
    \item[-] The lack of consensus on how the geometry derived from multiple stereo pairs should be refined or aggregated. Local point-wise operations are common to merge altitude values derived from different pairs, e.g. median \cite{dangelo2012dense, gong2019dsm, mari2019bundle} or k-medians \cite{facciolo2017automatic}. However, recent work has shown that deep learning approaches can greatly improve the result, e.g. by exploiting geometric priors related to urban areas \cite{bittner2018dsm, bittner2019late, liebel2020generalized, stucker2020resdepth}.
    \item[-] Very often, it is necessary to make adjustments or parameter tuning to handle different sources or types of satellite images \cite{ozcanli2015comparison, zhang2019leveraging}.
\end{itemize}

Neural rendering represents an opportunity to find a natural solution to the previous issues, as it automatically learns the optimal features and operations adapted to each individual 3D scene. The main advantage of traditional pipelines is preserved as no explicit geometry supervision is required: the learning is self-supervised and based solely on the color of the input images. This is a key difference with respect to other state-of-the-art deep learning methods for DSM generation from satellite imagery \cite{bittner2019late, stucker2020resdepth, gao2021rational, gomez2022experimental}, which depend on ground truth geometry models.

\vspace{0.5pt}

\subsection{Neural Radiance Fields}
\label{sec:nerf_fundamentals}

NeRF \cite{mildenhall2020nerf} represents a static scene as a continuous volumetric function $\mathcal{F}$, encoded by a fully-connected neural network. $\mathcal{F}$ predicts the emitted RGB color~$\mathbf{c} = (r, g, b)$ and a non-negative scalar volume density~$\sigma$ at a 3D point $\mathbf{x} = (x, y, z)$ of the scene seen from a viewing direction $\mathbf{d} = (d_x, d_y, d_z)$, i.e.
\vspace{-4pt}
\begin{equation}
    \mathcal{F}:(\mathbf{x}, \mathbf{d}) \mapsto (\mathbf{c}, \sigma).
    \label{eq:classic_nerf_inputs_outputs}
\end{equation}
Multi-view consistency is encouraged by restricting the network to predict the volume density $\sigma$ based only on the spatial coordinates $\mathbf{x}$, while allowing the color $\mathbf{c}$ to be predicted as a function of both $\mathbf{x}$ and the viewing direction $\mathbf{d}$. The dependency of $\mathbf{c}$ on the viewing direction allows to recreate specular reflections caused by to static light sources.

Given a set of input views and their camera poses, the training strategy is based on rendering the color of individual rays traced across the scene and projected onto the known pixels. Individual rays are chosen randomly, encouraging gradient flow at those ray intersections where the surface of the scene is susceptible of being located. Each ray $\textbf{r}$ is defined by a point of origin $\mathbf{o}$ and a direction vector $\mathbf{d}$. The color $\mathbf{c}(\mathbf{r})$ of a ray $\mathbf{r}(t) = \mathbf{o} + t\mathbf{d}$ is computed as
\vspace{-4pt}
\begin{equation}
    \mathbf{c}(\mathbf{r}) = \sum_{i=1}^{N}T_i\alpha_i\mathbf{c}_i.
    \label{eq:nerf_color_rendering}
\end{equation}
The rendered color $\mathbf{c}(\mathbf{r})$ results from the weighted integration of the colors $\mathbf{c}_i$ predicted at different points of the ray $\mathbf{r}$, which is discretized into $N$ 3D points $\mathbf{x}_i$ between the near and far bounds of the scene, $t_n$ and $t_f$. Each point $\mathbf{x}_i$ in $\mathbf{r}$ is obtained as $\mathbf{x}_i = \mathbf{o} + t_i\mathbf{d}$, where $t_i \in [t_n, t_f]$.
 
Following \eqref{eq:nerf_color_rendering}, the weight given to the color predicted for each point $\mathbf{x}_i$ of $\mathbf{r}$ is defined by a transmittance factor $T_i$ representing the probability that light reaches the point without hitting any other particle, and an alpha compositing value $\alpha_i$ encoding the opacity. Both $T_i$ and $\alpha_i$ are set according to the volume density $\sigma_i$ predicted for $\mathbf{x}_i$:
\vspace{-4pt}
\begin{equation}
    \alpha_i = 1 - \exp(-\sigma_i\delta_i);  \quad 
     T_i = \prod_{j=1}^{i-1} \left( 1 - \alpha_j \right),
     \label{eq:opacity_and_transmittance}
\end{equation}
where $\delta_i$ is the distance between two consecutive points along the ray, i.e. $\delta_i = t_{i+1} - t_i$. Higher values of $\sigma_i$ will result in larger opacity $\alpha_i$, indicating that $\mathbf{x}_i$ possibly belongs to a non-transparent surface. Occlusions are handled by the transmittance $T_i$, equal to the cumulative product of the inverse opacity. Even if $\mathbf{x}_i$ is given a large $\sigma_i$, $T_i$ only allows it to contribute decisively to the rendered color if it is not preceded by previous opaque points in the ray. 

Given \eqref{eq:opacity_and_transmittance}, the depth $d(\mathbf{r})$ observed in the direction of a ray $\mathbf{r}$ can be rendered in a similar manner to \eqref{eq:nerf_color_rendering} \cite{deng2021depth,roessle2021dense} as
\vspace{-4pt}
\begin{equation}
    d(\mathbf{r}) = \sum_{i=1}^{N}T_i\alpha_it_i.
    \label{eq:nerf_depth_rendering}
\end{equation}
NeRF is optimized by minimizing the mean squared error (MSE) between the rendered color and the real color of the input images, at the positions where the rays project:
\vspace{-2pt}
\begin{equation}
    \sum_{\mathbf{r} \in \mathcal{R}} \|  \mathbf{c}(\mathbf{r}) - \mathbf{c}_{\text{GT}}(\mathbf{r}) \|_2^2,
    \label{eq:nerf_classic_loss} 
\end{equation}
where $\mathbf{c}_{\text{GT}}(\mathbf{r})$ is the observed color of the pixel intersected by the ray $\mathbf{r}$, and $\mathbf{c}(\mathbf{r})$ is the color predicted by the NeRF using \eqref{eq:nerf_color_rendering}. $\mathcal{R}$ is the set of rays in each input batch.

\subsection{NeRF variants}

NeRF assumes that the density, radiance and illumination of the target 3D scene is constant. This is a strong limitation, as these conditions are rarely encountered outside laboratory settings. Many variants have been proposed to address this problem. In this section we briefly review three models that inspired our work.

\textbf{NeRF-W}~\cite{martin2021nerf} or \textit{NeRF in the Wild} gains robustness to radiometric variation and transient objects by learning to separate transient phenomena from the static scene. An extra head of fully-connected layers is used to predict a transient color $\mathbf{c}^{\tau}$ and volume density $\sigma^{\tau}$ for each input point, in addition to the usual $\mathbf{c}$ and $\sigma$. The transient outputs are linearly combined with the static ones to render the color of each ray. NeRF-W also uses the transient head to emit an uncertainty coefficient $\beta$, which measures the confidence of the network that a point belongs to a transient object. The value of $\beta$ is used in the loss function to reduce the impact of transient/unreliable pixels in the learning process.

\textbf{S-NeRF}~\cite{derksen2021shadow} or \textit{Shadow NeRF} is, to the best of our knowledge, the first attempt to apply NeRF for multi-view satellite photogrammetry. S-NeRF showed the benefits in geometry estimation of simultaneously exploiting the direction of solar rays to learn the amount of sunlight $s$ that reaches each point $\mathbf{x}$ of the scene. The direction of solar rays is a common metadata of satellite images. Our work can be seen as an extension of S-NeRF that incorporates a modeling of transient objects similar to \cite{martin2021nerf} and a representation of the camera models more adapted to satellite data.

\textbf{DS-NeRF}~\cite{deng2021depth} or \textit{Depth Supervised NeRF} incorporates a depth supervision term to the loss function to accelerate the learning and reduce the amount of input images. The depth supervision term exploits a sparse set of 3D points that belong to the surface of the scene, which can be easily retrieved using structure-from-motion (SfM) pipelines. SfM is a common pre-processing step in NeRF frameworks, as it can estimate the camera poses needed to cast the input rays. A similar strategy to DS-NeRF is used in \cite{roessle2021dense}, which converts the sparse point clouds into dense depth priors.

Other recent NeRF variants are yet to be investigated in the context of satellite imagery. Some works are focused on achieving smoother scene representations or reducing the number of input views: e.g. DietNeRF~\cite{jain2021putting} introduces an auxiliary semantic loss to maximize similarity between high-level features instead of RGB colors; Mip-NeRF~\cite{barron2021mipnerf} prevents blurring and aliasing in collections of images with different resolutions; PixelNeRF \cite{yu2021pixelnerf} describes a framework that is trained across multiple scenes and learns priors that can generalize to unseen scenes with few available images. Recent undergoing research is also progressing to extend NeRFs to dynamic scenes \cite{park2021nerfies, park2021hypernerf, pumarola2021d, xu2021h}, to gain efficiency and reduce the training time \cite{yang2021recursive, yu2021plenoctrees, hedman2021baking, muller2022instant} or to handle complex illumination settings, under arbitrary, multiple light sources \cite{bi2020neural, srinivasan2021nerv} or near-darkness conditions \cite{mildenhall2021nerf}.
\section{Method}
\label{sec:method}

Sat-NeRF represents the scene as a static surface with an albedo color, i.e. the intrinsic color of static objects. The model learns to predict the geometry and the albedo color simultaneously with a set of additional outputs, which seek to explain the transient phenomena observed in the input images without inducing changes in the scene geometry. 

We train the model following the ray casting strategy of NeRF (Section~\ref{sec:nerf_fundamentals}). Unlike the original NeRF~\eqref{eq:classic_nerf_inputs_outputs}, we assume a Lambertian surface and omit the color dependence on viewing angles. The inputs are
\begin{itemize}[noitemsep,topsep=0pt]
\setlength\itemsep{0.1em}
   \item[-] $\mathbf{x}$:~3-valued vector with the spatial coordinates of points located in the scene volume. $\mathbf{x}$ is part of a ray $\mathbf{r}$.
    \item[-] $\bm{\omega}$:~3-valued direction vector encoding the direction of solar rays. For each input image, $\bm{\omega}$ is extracted from the azimuth and elevation angles $(\theta, \phi)$ that indicate the position of the sun in the satellite image metadata.
    \item[-] $\mathbf{t}_j$:~$N^{(\text{t})}$-valued embedding vector, learned as a function of the image index $j$. The objective of $\mathbf{t}_j$ is to featurize the transient elements in the $j$-th view that cannot be explained by the position of the sun given by $\bm{\omega}$. We manually set $N^{(\text{t})}$\,$=$\,$4$.
\end{itemize}
The volumetric function of Sat-NeRF then writes \mbox{$\mathcal{F}:\!(\mathbf{x}, \bm{\omega}, \mathbf{t}_j)\mapsto(\sigma, \mathbf{c}_{\text{a}}, s, \mathbf{a}, \beta)$}, where the outputs are
\begin{itemize}[noitemsep,topsep=0pt]
\setlength\itemsep{0.1em}
    \item[-] $\sigma$:~scalar encoding the volume density at location $\mathbf{x}$.
    \item[-] $\mathbf{c}_{\text{a}}$:~albedo RGB color, which depends exclusively on the geometry, i.e. the spatial coordinates $\mathbf{x}$.
    \item[-] $s$:~shadow-aware shading scalar, learned as a function of $\mathbf{x}$ and the solar rays direction vector $\bm{\omega}$.
    \item[-] $\mathbf{a}$:~ambient RGB color, independent of scene geometry, that defines a global hue bias according to the position of the sun given by $\bm{\omega}$.
    \item[-] $\beta$:~uncertainty coefficient related to the probability that the color of $\mathbf{x}$ is explained by a transient object.
\end{itemize}

\subsection{Shadow-aware irradiance model}
\label{sec:irradiance_model}

\begin{figure*}[t]
 \begin{tabular}{c}
  \Tstrutsuper
  \input{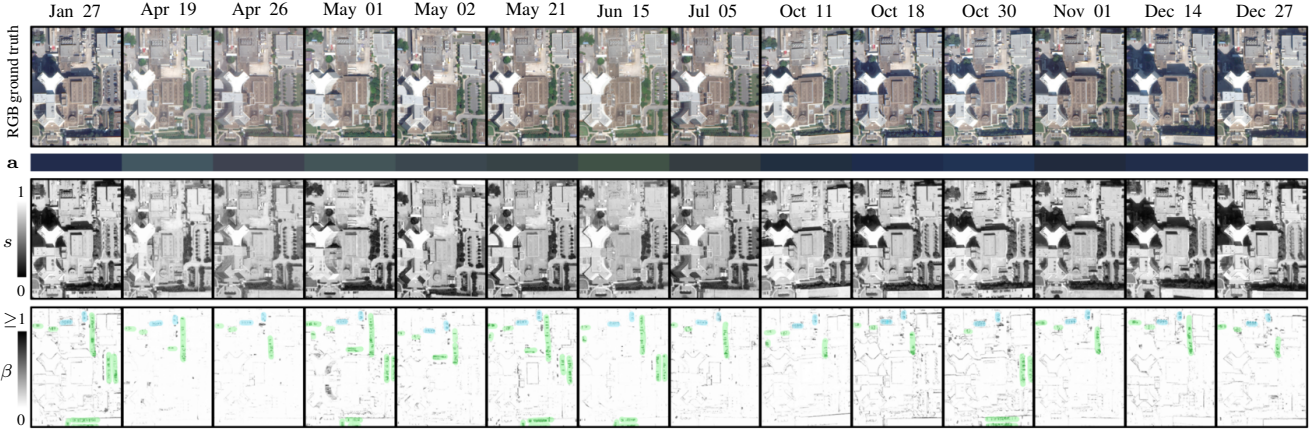}
  \end{tabular}
  \caption{The shading scalar $s$ related to the solar rays direction $\bm{\omega}$ learns shadows and material roughness. We observe that $\bm{\omega}$ is narrowly related to the acquisition date, causing the ambient color $\mathbf{a}$ associated with the low values of $s$ (see \eqref{eq:irradiance_model}) to capture a mixture of phenomena, including seasonal changes reflected in the vegetation. The uncertainty prediction $\beta$ does not affect shadows and concentrates on small color inconsistencies, mostly caused by cars in open-air parkings (green marks), large fans in rooftops (blue marks) or building edges.}
    \label{fig:shading_study}
\end{figure*}

This section describes how Sat-NeRF predicts the color $\mathbf{c}(\mathbf{r})$ of a ray $\mathbf{r}$ projected onto a certain pixel. We keep the rendering as in \eqref{eq:nerf_color_rendering} and \eqref{eq:nerf_depth_rendering}, with the transmittance and opacity factors as defined in \eqref{eq:opacity_and_transmittance}, but adopt the shadow-aware irradiance model proposed in S-NeRF \cite{derksen2021shadow} to compute the color $\mathbf{c}$ at each point $\mathbf{x}$ of a ray $\mathbf{r}$: \vspace{-3pt}
\begin{equation}
    \mathbf{c}(\mathbf{x}, \bm{\omega}, \mathbf{t}_j) = \mathbf{c}_{\text{a}}(\mathbf{x}) \cdot \big( s(\mathbf{x}, \bm{\omega}) + (1 - s(\mathbf{x}, \bm{\omega})) \cdot \mathbf{a}(\bm{\omega}) \big),
    \label{eq:irradiance_model}\vspace{-3pt}
\end{equation}
where $\mathbf{c}(\mathbf{x}, \bm{\omega}, \mathbf{t}_j)$ substitutes $\mathbf{c}_i$ in \eqref{eq:nerf_color_rendering}. The shading scalar $s(\mathbf{x}, \bm{\omega})$ takes values between 0 and 1 and is used to add shadows by darkening the albedo (Figure~\ref{fig:shading_study}). Ideally, $s\approx$\,1 in those 3D points directly illuminated by the sun, whose color should be entirely explained by the albedo $\mathbf{c}_{\text{a}}(\mathbf{x})$.

In addition, \eqref{eq:irradiance_model} attempts to capture the bluish hues of shadows \cite{arevalo2008shadow, ma2008shadow} by means of the ambient color $\mathbf{a}(\bm{\omega})$, which contributes to the points where $s$ takes values closer to 0. In practice, we find that the direction of the solar rays $\bm{\omega}$ is narrowly related to the acquisition date (especially if the satellite passes at the same hours of the day), as shown in Figure~\ref{fig:shading_study}. Thus, $\mathbf{a}(\bm{\omega})$ ends up capturing ambient irradiance  due to a mixture of phenomena, which is related to $\bm{\omega}$ but also date-specific conditions like weather or seasonal changes. 

\captionsetup{skip=0pt}
\captionsetup{skip=8pt}

As observed in S-NeRF \cite{derksen2021shadow}, the shading scalar $s(\mathbf{x}, \bm{\omega})$ in \eqref{eq:irradiance_model} can produce unrealistic results for solar rays directions that are not seen in the training data. This can be minimized by adding a \textit{solar correction} term to the loss:
\begin{equation}
    L_{\text{SC}}(\mathcal{R}_{\text{SC}}) = \! \sum\limits_{\mathbf{r} \in \mathcal{R}_{\text{SC}}} \left( \sum\limits_{i=1}^{N_{\text{SC}}}(T_i - s_i)^2 + 1 - \sum\limits_{i=1}^{N_{\text{SC}}}T_i\alpha_is_i \right),
    \label{eq:snerf_solarcorrection}
\end{equation}
where $\mathcal{R}_{\text{SC}}$ is a secondary batch of solar correction rays. Note that the rays in $\mathcal{R}_{\text{SC}}$ follow the direction of solar rays $\bm{\omega}$, while the rays in $\mathcal{R}$, in the main term of the loss \eqref{eq:nerf_classic_loss}, follow the viewing direction of the camera. 

The solar correction term \eqref{eq:snerf_solarcorrection} uses the learned geometry, encoded by the transmittance $T_i$ and opacity $\alpha_i$~\eqref{eq:opacity_and_transmittance}, to further supervise the learning of the shadow-aware shading $s(\mathbf{x}, \bm{\omega})$. The first part of \eqref{eq:snerf_solarcorrection} enforces that, for each ray $\mathbf{r}$ in $\mathcal{R}_{\text{SC}}$, the $s_i$ predicted at the $i$-th point should resemble $T_i$, i.e. high values before reaching the visible surface, low values afterwards (both $s_i$ and $T_i$ take values between \mbox{0 and 1)}. The second part of \eqref{eq:snerf_solarcorrection} encourages that the integration of $s$ over $\mathbf{r}$ reaches 1, since non-occluded and non-shadow areas have to be mostly explained by the albedo in \eqref{eq:irradiance_model}.

\subsection{Uncertainty weighting for transient objects}
\label{sec:transient_objects}

Similarly to NeRF-W~\cite{martin2021nerf}, we use the task-uncertainty learning approach introduced in \cite{kendall2018multi} to gain robustness to transient objects by means of $\beta$. In our context, transient objects are punctual local changes across the input images that cannot be explained by the static surface or the available metadata, like the position of the sun. The irradiance model~\eqref{eq:irradiance_model} does not handle transient objects explicitly. As a result, we observe that $s$ and $\sigma$ usually try to account for them, leading to wrong depth predictions, as shown in Figure~\ref{fig:beta_study}. Thanks to $\beta$, Sat-NeRF is given some margin to ignore the color inconsistencies caused by these objects.

The uncertainty prediction $\beta$ weights the contribution of each ray to the MSE between rendered and known colors:
\begin{equation}
    L_{\text{RGB}}(\mathcal{R}) \!= \!\sum_{\mathbf{r} \in \mathcal{R}} \frac{\| \mathbf{c}(\mathbf{r}) - \mathbf{c}_{\text{GT}}(\mathbf{r}) \|_2^2}{2\beta'(\mathbf{r})^2} + \left( \frac{\log\beta'(\mathbf{r}) \!+ \!\eta}{2} \right),
    \label{eq:satnerf_loss_part1}
\end{equation}
where $\beta'(\mathbf{r}) = \beta(\mathbf{r}) + \beta_{\text{min}}$. In \eqref{eq:satnerf_loss_part1}, we use $\beta_{\text{min}} = 0.05$ and $\eta = 3$ to avoid negative values in the logarithm. The role of the logarithm in $L_{\text{RGB}}$ is to prevent $\beta$ from converging to infinity to solve the problem. In this way the model is forced to find a compromise between the uncertainty coefficients $\beta$ and the differences of colors.  

The $\beta(\mathbf{r})$ associated with a ray $\mathbf{r}$ is obtained by integrating the uncertainty predictions across the $N$ points of $\mathbf{r}$:
\vspace{-3pt}
\begin{equation}
    \beta(\mathbf{r}) = \sum_{i=1}^{N}T_i\alpha_i\beta(\mathbf{x}_i, \mathbf{t}_j),\vspace{-3pt}
\end{equation}%
where $\beta(\mathbf{x}_i, \mathbf{t}_j)$, is the uncertainty coefficient predicted at the $i$-th point of $\mathbf{r}$. Sat-NeRF learns to predict the uncertainty $\beta$ at each point of the scene based on its spatial coordinates $\mathbf{x}$ (some areas are more likely to exhibit transient objects, e.g. open-air parkings in Figure~\ref{fig:shading_study}) and on the transient embedding vector $\mathbf{t}_j$ of each input training image. Depending on each view, the areas typically affected by transient objects will arbitrarily differ to a greater or lesser extent with respect to the albedo. Note that the embedding vector $\mathbf{t}_j$ is learned from the image index $j$ during training.\footnote{At test time, we use an arbitrary $\mathbf{t}_j$ selected from the training set.}

We find that it is better to start using $\beta$ after the second epoch, when the shadow-aware shading $s$ is already well initialized. Otherwise the model may use $\beta$ to overlook shadow areas instead of trying to explain them with $s$. Thus, we replace \eqref{eq:satnerf_loss_part1} with \eqref{eq:nerf_classic_loss} in the first two epochs.

\subsection{Point sampling from satellite RPC models}
\label{sec:ray_sampling}

Sat-NeRF casts rays directly using the RPC camera models of a set of satellite images. The RPC model is widely used for optical satellite imagery, as it allows to describe complex acquisition systems independently of satellite-specific physical modeling~\cite{grodecki2001ikonos, akiki2021robust}. Each RPC is defined by a projection function (to project 3D points onto image pixels) and its inverse, the localization function.

The use of RPCs in a NeRF framework represents an improvement with respect to previous work with satellite data. In \mbox{S-NeRF} \cite{derksen2021shadow} the RPC model of each input view is replaced with a custom simplified pinhole camera matrix, which is the common representation used in NeRF for close-range imagery~\cite{mildenhall2020nerf}. The RPC-based sampling described here corresponds to a more general approach, which also leads to better results (see Section~\ref{sec:evaluation}).

We denote the minimum and maximum altitudes of the scene as $h_{\text{min}}$ and $h_{\text{max}}$, respectively.\footnote{The altitude bounds $[h_{\text{min}}$, $h_{\text{max}}]$ can be selected in various ways, e.g. from a large-scale elevation model extracted from low-resolution data.} The ray that crosses the scene and intersects the pixel $\mathbf{p}$ of the $j$-th image is modeled as a straight line between an initial and a final 3D point, i.e. $\mathbf{x}_\text{start}$ and $\mathbf{x}_\text{end}$. These boundary points are obtained by localizing the pixel $\mathbf{p}$ at $h_{\text{min}}$ and $h_{\text{max}}$, using the RPC localization function $\mathcal{L}_j$ of the $j$-th image:
\begin{equation}
    \mathbf{x}_\text{start} =  \mathcal{L}_j(\mathbf{p}, h_{\text{max}})_{\text{ECEF}}; \quad
    \mathbf{x}_\text{end} = \mathcal{L}_j(\mathbf{p}, h_{\text{min}})_{\text{ECEF}}, \label{eq:ray_boundaries}
\end{equation}
where the subindex ECEF indicates that the 3D points returned by the localization function $\mathcal{L}_j$ are converted to the Earth-centered, Earth-fixed coordinate system (or geocentric system), to work in a cartestian system of reference.

Given $\mathbf{x}_\text{start}$  and $\mathbf{x}_\text{end}$, the origin $\mathbf{o}$ and direction vector $\mathbf{d}$ of the ray $\mathbf{r}(t) = \mathbf{o} + t\mathbf{d}$ that intersects the pixel $\mathbf{p}$ of the $j$-th image are expressed as 
\begin{equation}
     \mathbf{o} = \mathbf{x}_\text{start};  \quad 
    \mathbf{d} = \frac{\mathbf{x}_\text{end} - \mathbf{x}_\text{start}}{\| \mathbf{x}_\text{end} - \mathbf{x}_\text{start} \|_2}. \label{eq:ray_origin_direction}
\end{equation}
The point of maximum altitude, $\mathbf{x}_\text{start}$, which is the closest to the camera, is taken as the origin $\mathbf{o}$ of the ray. The boundaries of the ray $\mathbf{r}(t) = \mathbf{o} + t\mathbf{d}$, i.e. $[t_{\text{min}}, t_{\text{max}}]$, are set as $t_{\text{min}} =0$ and $t_{\text{max}} = \| \mathbf{x}_\text{end} - \mathbf{x}_\text{start} \|_2$. Since working with ECEF coordinates is impractical, due to large values used in the representation, we normalize all ray points in the interval $[-1, 1]$ using an offset subtraction and scaling procedure similar to the one used in the RPC functions~\cite{grodecki2001ikonos}. The set of 3D points resulting from localizing all pixels in the input images at $h_{\text{min}}$ and $h_{\text{max}}$ is used to compute the offset and scale in each spatial dimension.

\subsection{RPC refinement for improved performance}
\label{sec:depth_supervision}

Bundle adjustment approaches are a common good practice in remote sensing to correct inconsistencies between a collection of RPC models observing the same scene \cite{grodecki2003block, ozcanli2014automatic, mari2019bundle, mari2021generic}. In particular, bundle adjustment methods correct the RPCs by minimizing the reprojection error of a set of corresponding points seen across the images~\cite{triggs1999bundle}.

In absence of a prior RPC refinement, a 3D point projected with different raw RPC functions often falls on non-coincident image points, by a distance of up to tens of pixels~\cite{grodecki2003block}. This would cause a systematic loss of accuracy in any NeRF methodology for satellite imagery, because rays traced from corresponding pixels of different views would not intersect at an exact point in the object space. To prevent this situation, before training Sat-NeRF, we apply the bundle adjustment method described in \cite{mari2021generic}, which performs a relative correction of the RPC models of all input images. The set of points used by the bundle adjustment is derived from correspondences of SIFT keypoints \cite{lowe2004distinctive}.

While the refined RPC models directly increase the accuracy of the point sampling strategy described in Section~\ref{sec:ray_sampling}, a prior bundle adjustment can also improve the Sat-NeRF performance in other ways. \mbox{DS-NeRF}~\cite{deng2021depth} discussed how the training of a NeRF can benefit from a sparse set of previously known 3D points, under the idea that such points can be easily produced using SfM pipelines. In the case of satellite imagery, the bundle adjustment produces an equivalent set of sparse 3D points derived from image features \cite{broxton20093d, shean2016automated, pan2021self, mari2021generic}. Based on this idea, we explore the benefits of adding the depth supervision term proposed in \cite{deng2021depth} to the loss of our Sat-NeRF model:
\begin{equation}
    L_{\text{DS}}(\mathcal{R}_{\text{DS}}) = \sum_{\mathbf{r} \in \mathcal{R}_{\text{DS}}} w(\mathbf{r}) \left( d(\mathbf{r}) - \| \mathbf{X}(\mathbf{r}) - \mathbf{o}(\mathbf{r}) \|_2 \right)^2,
    \label{eq:depth_supervision}
\end{equation}
where $d(\mathbf{r})$ is the depth \eqref{eq:nerf_depth_rendering} predicted for a ray $\mathbf{r}$, whose origin point is $\mathbf{o}(\mathbf{r})$. If $\mathbf{r}$ intersects $\mathbf{X}(\mathbf{r})$, a known 3D point, then $\| \mathbf{X}(\mathbf{r}) - \mathbf{o}(\mathbf{r}) \|_2$ is the target depth to be learned. $\mathcal{R}_{\text{DS}}$ denotes a batch of rays that intersects known 3D points. Since the pixel coordinates associated with these 3D points are already provided by the bundle adjustment, all rays in $\mathcal{R}_{\text{DS}}$ can be defined as explained in Section~\ref{sec:ray_sampling}. Normalized coordinates between $[-1, 1]$ are used in \eqref{eq:depth_supervision} to represent points in the object space, for consistency with Section~\ref{sec:ray_sampling}.

Similarly to \cite{deng2021depth}, we only use $L_{\text{DS}}$ in the initial 25\% of training iterations. In our experience, this proportion is usually enough to gain accuracy in the learned geometry. Observe that the contribution of each depth supervision ray $\mathbf{r}$ in $\mathcal{R}_{\text{DS}}$ is weighted by $w(\mathbf{r})$ in \eqref{eq:depth_supervision}, where $w(\mathbf{r})$ is a scalar set according to the reprojection error of each point $\mathbf{X}(\mathbf{r})$ provided by the bundle adjustment.

\subsection{Multi-task loss and network architecture}
\label{sec:architecture}

The main term of the Sat-NeRF loss function is the $L_{\text{RGB}}$ defined in \eqref{eq:satnerf_loss_part1}, which is complemented by the solar correction term $L_{\text{SC}}$ \eqref{eq:snerf_solarcorrection} and the depth supervision term $L_{\text{DS}}$ \eqref{eq:depth_supervision}. The complete loss function can be expressed as
\begin{equation}
    L = L_{\text{RGB}}(\mathcal{R}) + \lambda_{\text{SC}}L_{\text{SC}}(\mathcal{R}_{\text{SC}}) + \lambda_{\text{DS}}L_{\text{DS}}(\mathcal{R}_{\text{DS}}),
\end{equation}
where $\lambda_{\text{SC}}$ and $\lambda_{\text{DS}}$ are an arbitrary weight given to each secondary term. We empirically find $\lambda_{\text{SC}}\,=\,0.1/3$ and $\lambda_{\text{DS}}\,=\,1000/3$ to provide good results, to keep the secondary terms sufficiently relevant but below the magnitude of $L_{\text{RGB}}$. For depth supervision, we used \mbox{$\sim$2k-10k} bundle adjustment points depending on the output of \cite{mari2021generic} for each area of interest. $\mathcal{R}$, $\mathcal{R}_{\text{SC}}$ and $\mathcal{R}_{\text{DS}}$ have the same batch size.

\begin{figure}[t]
  \begin{tabular}{c}
  \Tstrutarch
  \input{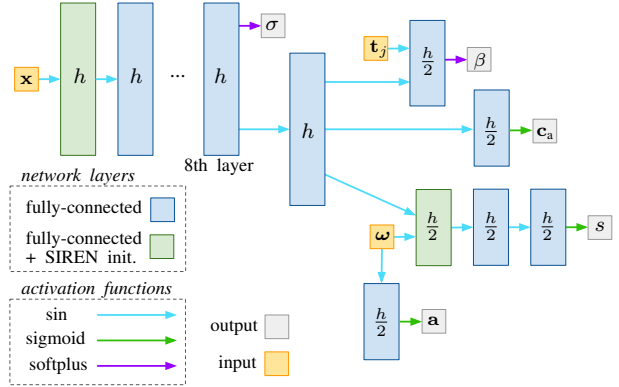}
  \end{tabular}
  \caption{Sat-NeRF network architecture, where $\mathbf{x}$ are the input spatial coordinates, $\bm{\omega}$ is the direction of solar rays and $\mathbf{t}_j$ is the learned transient embedding of image $j$. The model predicts the volume density $\sigma$, the components of the irradiance model \eqref{eq:irradiance_model}, i.e. albedo color $\mathbf{c}_{\text{a}}$, shading scalar $s$, ambient color $\mathbf{a}$, and an uncertainty coefficient $\beta$ to weight the impact of transient objects.}
    \label{fig:architecture}
\end{figure}

The architecture of Sat-NeRF is shown in Figure~\ref{fig:architecture}. The main block of fully-connected layers, with $h$ channels per layer, is dedicated to the prediction of the static properties of the scene: the volume density $\sigma$ and the albedo color $\mathbf{c}_{\text{a}}$. A secondary head is added with fewer layers and half as many channels per layer to estimate the shading scalar $s$ based on the direction of solar rays $\bm{\omega}$ and the vector of $h$ geometry-related features learned by the main block. Lastly, two single-layer heads are used to predict the uncertainty coefficient $\beta$ and the ambient color $\mathbf{a}$, from the transient embedding vector $\mathbf{t}_j$ and $\bm{\omega}$, respectively.

We employ SIREN layers with the initialization proposed in \cite{sitzmann2020implicit}, as suggested in \cite{derksen2021shadow}. The use of a softplus function to predict $\sigma$ was crucial to achieve satisfactory results. The uncertainty $\beta$ is also produced by a softplus \cite{martin2021nerf}, which yields a smoother optimization problem compared to the usual ReLU \cite{barron2021mipnerf}. The rest of the outputs result from sigmoid functions, since they are related to normalized RGB values and have to be in the interval $[0, 1]$. The value of $h$ should be adjusted according to the resolution and the size of the observed area. In this work we set $h = 512$.
\section{Evaluation}
\label{sec:evaluation}

We evaluate Sat-NeRF on different areas of interest (AOI) of the \textit{2019 IEEE GRSS Data Fusion Contest} \cite{bosch2019semantic, le20192019}, which provides 26 Maxar WorldView-3 images collected between 2014 and 2016 over the city of Jacksonville, Florida, US. From this data, we take as input a set of RGB crops of varying size, around $800 \! \times \! 800$ pixels, with a resolution of 0.3 m/pixel at nadir, covering $256 \! \times \! 256$ m for each AOI. The indices of the selected AOIs and the number of training and test images that were used are listed in Table~\ref{table:dataset_description}.

\begin{table}[t]
    \centering \footnotesize{
  \begin{tabular}{| c@{\hskip 0.24cm}|@{\hskip 0.24cm}c@{\hskip 0.24cm}|@{\hskip 0.24cm}c@{\hskip 0.24cm}|@{\hskip 0.24cm}c@{\hskip 0.24cm}|@{\hskip 0.24cm}c@{\hskip 0.24cm}|}
  \hline
  Area index & 004 & 068 & 214 & 260 \\ \hline
  \# train/test & 9/2 & 17/2 & 21/3 & 15/2 \\ \hline
  Alt. bounds [m] & {\footnotesize [-24, 1]} & {\footnotesize [-27, 30]} & {\footnotesize [-29, 73]} & {\footnotesize [-30, 13]} \\ \hline
  \end{tabular} }
    \caption{Number of training and test images used for each area, and the altitude bounds of the scene considered in each case.}
    \label{table:dataset_description}
\end{table}

In all conducted experiments we use a single NeRF model, trained with an Adam optimizer starting with a learning rate of $5e^{-4}$, which is decreased at every epoch by a factor $\gamma=0.9$ according to a step scheduler. The batch size is 1024 rays, and each ray $\mathbf{r}$ is discretized into 64 uniformly distributed 3D points. Training takes 300k iterations to converge, resulting in $\sim$10\,h if a single batch of rays is used at each training iteration, or $\sim$20\,h if a secondary term for solar correction or depth supervision is added to the loss (trained on a GPU with 16~GB RAM). We used bundle adjusted RPCs (Section~\ref{sec:depth_supervision}) unless otherwise noted.

\subsection{Ablation study}
\label{sec:ablation_study}

We evaluated the Sat-NeRF model starting from a simple NeRF and gradually adding new components. To this end, we propose three categories of experiments that are discussed below. Table~\ref{table:results} shows the quantitative results.

{\bf{Category 1.}} Rows 0-3 are an ablation study dedicated to the irradiance model and the solar correction term described in Section~\ref{sec:irradiance_model}. We verify that the S-NeRF irradiance model outperforms a basic NeRF and is strengthened by the solar correction term. Comparing our results with the ones reported in the original S-NeRF work \cite{derksen2021shadow} reveals the impact of the proposed RPC-based point sampling and bundle adjustment detailed in Section~\ref{sec:method}, to which we attribute the difference between the metrics of row 0 and row 3. 

{\bf{Category 2.}} Rows 4-5 assess our Sat-NeRF model, which incorporates the uncertainty prediction of $\beta$ and employs \eqref{eq:satnerf_loss_part1} as main term of the loss function. These rows show that the uncertainty modeling improves both the learned geometry and the novel view synthesis, as illustrated in Figure~\ref{fig:beta_study}.   Compared to the best S-NeRF results (row 3), Sat-NeRF (row 4) provides higher PSNR/SSIM and similar or even smaller altitude MAE without requiring a solar correction term. This insight could be exploited in settings that cannot afford additional training time to process a secondary batch of rays for solar correction. If we add the solar correction term \eqref{eq:snerf_solarcorrection} to the Sat-NeRF loss \eqref{eq:satnerf_loss_part1}, the altitude MAE decreases even more: row 5 outperforms all the previous configurations across all AOIs. 

{\bf{Category 3.}} Rows 6-8 demonstrate the benefits of using a prior bundle adjustment to refine the RPC models of the satellite images (Section~\ref{sec:depth_supervision}). The comparison of rows 6 and 5 reveals that the use of unrefined RPCs induces a performance drop: both PSNR/SSIM and altitude MAE are worse. Lastly, \mbox{rows 7-8} add the depth supervision term \eqref{eq:depth_supervision} to the Sat-NeRF loss \eqref{eq:satnerf_loss_part1} with and without solar correction, to leverage the sparse point cloud provided by the bundle adjustment. This strategy proves to be beneficial to further improve the altitude MAE in some areas (214 and 260). In other areas (004 and 068), the sparse point cloud may contain outliers, especially in multi-date collections, leading to slightly worse MAE compared to a plain Sat-NeRF loss.

\captionsetup{skip=9pt}
\captionsetup{belowskip=10pt}
\input{fig_beta_study}

\captionsetup{skip=5pt}
\input{fig_dsms}
\captionsetup{skip=10pt}

\captionsetup{belowskip=3pt}
\newcommand{\nb}[1]{{\footnotesize #1}}
\newcommand{\customfont}[1]{{\scriptsize #1}}

\definecolor{amber}{rgb}{1.0, 0.75, 0.0}
\newcommand{\goldmedal}{%
    \protect\tikz[]{
        \fill[amber] (0,0) circle (0.09);
    }
}

\definecolor{silver}{rgb}{0.75, 0.75, 0.75}
\newcommand{\silvermedal}{%
    \protect\tikz[]{
        \fill[silver] (0,0) circle (0.09);
    }
}

\definecolor{bronze}{rgb}{0.8, 0.5, 0.2}
\newcommand{\bronzemedal}{%
    \protect\tikz[]{
        \fill[bronze] (0,0) circle (0.09);
    }
}

\newcommand{\nomedal}{%
    \protect\tikz[]{
        \fill[white] (0,0) circle (0.09);
    }
}

\begin{table*}[t]
 \centering \small{
    \begin{tabular}{ l| c c c c | c c c c | c c c c }
  \hline %
  \multicolumn{1}{c|}{\Tstrutsmall} & \multicolumn{4}{c|}{\textbf{PSNR} 	$\bm{\uparrow}$} & \multicolumn{4}{c|}{\textbf{SSIM} $\bm{\uparrow}$} & \multicolumn{4}{c}{\textbf{Altitude MAE [m] $\bm{\downarrow}$}} \\
  \multicolumn{1}{c|}{Area index}& 004 & 068 & 214 & 260 & 004 & 068 & 214 & 260 &
                      004 \nomedal & 068 \nomedal & 214 \nomedal & 260 \nomedal \\
                      \Cline{1.2pt}{1-13}%
  \footnotesize{0. S-NeRF + SC \cite{derksen2021shadow}} & 
   \nb{---} & \nb{---} & \nb{---} & \nb{---} & 
   \nb{0.344} & \nb{0.459} & \nb{0.384} & \nb{0.416} &
   \nb{4.418} \nomedal & \nb{3.644} \nomedal  & \nb{4.829} \nomedal & \nb{7.173} \nomedal \\ \cline{1-13}
   \footnotesize{1. NeRF} & 
   \nb{17.93} & \nb{10.26} & \nb{15.26} & \nb{14.95} & 
   \nb{0.559} & \nb{0.536} & \nb{0.736} & \nb{0.443} & 
   \nb{3.327} \nomedal & \nb{2.591} \nomedal & \nb{2.691} \nomedal & \nb{3.257} \nomedal \\ \cline{1-13}
   \footnotesize{2. S-NeRF} & 
   \nb{25.87} & \nb{24.20} & \nb{24.51} & \nb{21.52} & 
   \nb{0.864} & \nb{0.900} & \nb{0.939} & \nb{0.829} & 
   \nb{1.830} \nomedal & \nb{1.496} \nomedal & \nb{3.687} \nomedal & \nb{3.245} \nomedal\\ \cline{1-13}
  \footnotesize{3. S-NeRF + SC} & 
   \nb{26.14} & \nb{24.07} & \nb{24.93} & \nb{21.24} & 
   \nb{0.871} & \nb{0.891} & \nb{0.943} & \nb{0.825} & 
   \nb{1.472} \nomedal & \nb{1.374} \nomedal & \nb{2.406} \nomedal & \nb{2.299} \nomedal\\ \Cline{1.2pt}{1-13}%
  \footnotesize{4. Sat-NeRF} &    
   \nb{26.16} & \nb{24.80} & \nb{25.54} & \nb{21.88} & 
   \nb{0.876} & \nb{0.903} & \nb{0.951} & \nb{0.840} & 
   \nb{1.416} \nomedal & \nb{1.275} \bronzemedal & \nb{2.125} \nomedal  & \nb{2.428} \nomedal \\ \cline{1-13}
  \footnotesize{5. Sat-NeRF + SC} &   
   \nb{26.67} & \nb{25.07} & \nb{25.50} & \nb{21.78} & 
   \nb{0.884} & \nb{0.908} & \nb{0.950} & \nb{0.842} & 
   \textbf{\nb{1.288}} \goldmedal & \textbf{\nb{1.249}} \silvermedal & \nb{2.009} \nomedal & \nb{1.864} \nomedal \\ \Cline{1.2pt}{1-13}%
  \footnotesize{6. Sat-NeRF + SC (no BA)} &
   \nb{21.55} & \nb{22.87} & \nb{24.53} & \nb{20.96} & 
   \nb{0.571} & \nb{0.874} & \nb{0.942} & \nb{0.816} & 
   \nb{1.577} \nomedal & \nb{1.392} \nomedal & \nb{2.176} \nomedal  & \nb{1.875} \nomedal \\ \cline{1-13}
  \footnotesize{7. Sat-NeRF + DS} &
   \nb{26.43} & \nb{25.27} & \nb{25.69} & \nb{21.94} & 
   \nb{0.879} & \nb{0.913} & \nb{0.952} & \nb{0.842} & 
   \nb{1.420} \nomedal  & \nb{1.298} \nomedal & \nb{1.714} \silvermedal & \textbf{\nb{1.624}} \goldmedal \\ \cline{1-13}
     \footnotesize{8. Sat-NeRF + DS + SC} &
   \nb{26.62} & \nb{25.00} & \nb{25.66} & \nb{21.66} & 
   \nb{0.881} & \nb{0.909} & \nb{0.952} & \nb{0.839} & 
   \nb{1.366} \silvermedal  & \nb{1.277} \nomedal & \textbf{\nb{1.676}} \goldmedal & \nb{1.638} \silvermedal \\ \Cline{1.2pt}{1-13}%
  \footnotesize{ \, \, S2P (10 pairs) \cite{facciolo2017automatic}} &
   \nb{---} & \nb{---} & \nb{---} & \nb{---} & 
   \nb{---} & \nb{---} & \nb{---} & \nb{---} & 
   \nb{1.370} \bronzemedal & \nb{1.174} \goldmedal & \nb{1.811} \bronzemedal & \nb{1.640} \bronzemedal \\ \hline %
  \end{tabular}}
    \caption{Numerical results using the test images (unseen in training). Rows 1-8 use the RPC-based sampling introduced in Section~\ref{sec:ray_sampling}. $\lambda_{\text{SC}}$\,$=$\,0.05/3 in row 3, otherwise $\lambda_{\text{SC}}$\,$=$\,0.1/3 and $\lambda_{\text{DS}}$\,$=$\,1000/3 where solar correction (SC) and depth supervision (DS) are used. The best altitude MAE values are given \mbox{gold~\goldmedal}, \mbox{silver~\silvermedal} and \mbox{bronze~\bronzemedal} medals. Sat-NeRF + SC and/or DS are the most awarded NeRF variants.}
    \label{table:results}
\end{table*}

\subsection{Comparison to traditional stereo pipelines}

Sat-NeRF learns high quality 3D models, similar in accuracy to those obtained with satellite stereo pipelines relying on traditional algorithms for stereo matching~\cite{beyer2018ames, defranchis2014automatic}. In this work, we compare the DSMs produced by Sat-NeRF with a multi-view stereo DSM of the same area generated with S2P~\cite{defranchis2014automatic, facciolo2017automatic}, the satellite stereo pipeline that won the \textit{2016 IARPA Multi-View Stereo 3D Mapping Challenge}~\cite{bosch2016multiple}.

We follow the methodology described in \cite{facciolo2017automatic} to produce the S2P DSMs. For each AOI, we manually select 10 stereo pairs for disparity estimation. The selection criterion prioritizes pairs with an angle between views of 5 to 45 degrees, with a maximum incidence angle of 40 degrees for each view. Within this set, we take the 10 pairs with closer acquisition dates and run S2P. The RPCs used by S2P were the same used to train Sat-NeRF, i.e. all RPCs are bundle adjusted using \cite{mari2021generic}. The 10 pairwise models are fused into a single DSM by taking the median altitude at each cell. To maximize the quality of the S2P DSMs we used the panchromatic product of the WorldView-3 images, instead of the RGB crops employed to train Sat-NeRF. Considering that the RGB images have a compressed dynamic (integer values in $[0,255]$), i.e with less texture and more saturated areas, the Sat-NeRF DSMs are very encouraging compared to the state of the art with manual pair selection.

\begin{figure}[t]
\begin{tabular}{c}
 \begin{tikzpicture}
  \node[anchor=south west, inner sep=0] (img) at (0,0){
  \includegraphics[width=8cm]{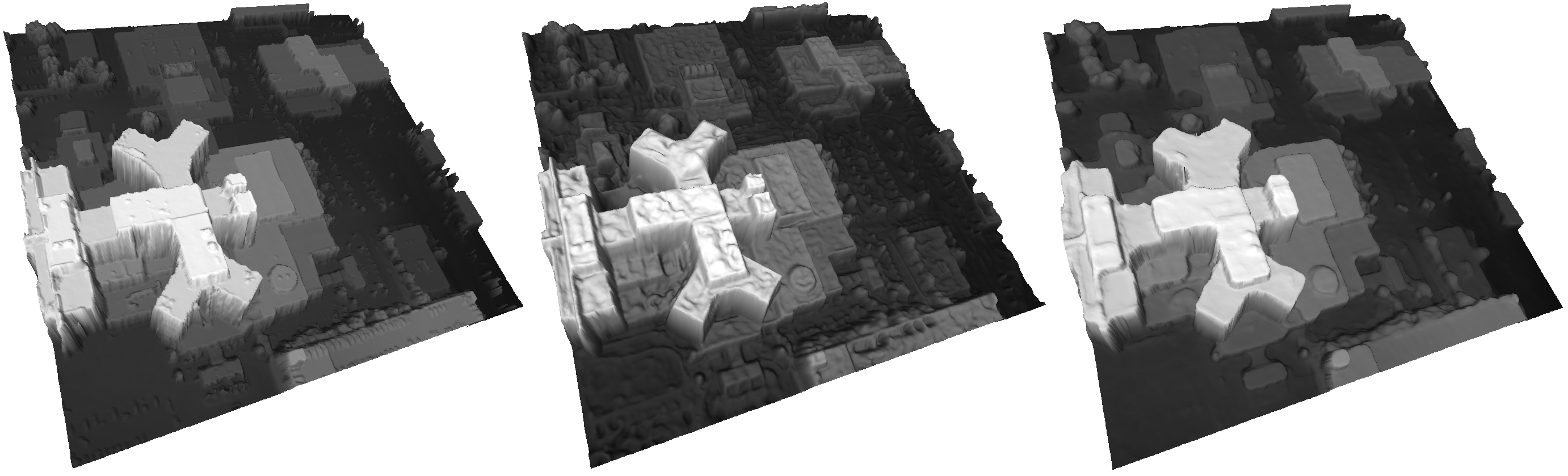}};
  \begin{scope}[x={(img.south east)},y={(img.north west)}]
    \node[text width=1.7cm]  at (0.33,0.15) {\scriptsize Lidar};
    \node[text width=2cm]  at (0.665,0.15) {\scriptsize Sat-NeRF};
    \node[text width=1.7cm]  at (0.99,0.15) {\scriptsize S2P};
  \end{scope}
  \end{tikzpicture}
  \end{tabular}
  \caption{3D visualization of the lidar, Sat-NeRF and S2P DSMs shown in Figure~\ref{fig:results_detail} (068). Compared to S2P, Sat-NeRF provides finer details and sharper edges but exhibits local irregularities.}
    \label{fig:surface_flatness_study}
\end{figure}

As shown in Figure~\ref{fig:results_detail} and~\ref{fig:surface_flatness_study}, structures are more detailed in Sat-NeRF DSMs, but S2P provides more regular surfaces. Numerically, the global altitude MAE obtained with Sat-NeRF can be slightly better compared to the S2P DSMs (Table~\ref{table:results}, last row), which are affected by single-point outliers. Future work points to hybrid methods or the aggregation of contour-preserving regularization techniques.

\captionsetup{skip=5pt}
\captionsetup{skip=9pt}
\section{Conclusion}
\label{sec:conclusion}

We introduced Sat-NeRF, a NeRF variant adapted for multi-date collections of multi-view satellite images. The geometry and appearance of permanent structures are simultaneously learned using a main backbone, while shadows and transient objects are learned by secondary heads.

The proposed method achieves state-of-the-art results in novel view synthesis and 3D modeling from satellite imagery. It also highlights the benefits of incorporating well-known techniques for satellite image processing into a NeRF framework. In particular, we show how to represent the input cameras using the RPC models characteristic of satellite images, instead of the pinhole cameras commonly used in NeRF for close-range imagery. We also demonstrate the advantages of applying a bundle adjustment step before training time, to improve reconstruction quality and, optionally, to provide additional cues for depth supervision.

\section{Acknowledgements}
This work was supported by a grant from Région Île-de-France. It was also partly financed by Office of Naval research grant N00014-17-1-2552, MENRT, and Kayrros. This work was performed using HPC resources from GENCI–IDRIS (grants 2021-AD011012453 and 2022-AD011011801R1) and from the “Mésocentre” computing center of CentraleSupélec and ENS Paris-Saclay supported by CNRS and Région Île-de-France (\href{http://mesocentre.centralesupelec.fr}{\tt\small http://mesocentre.centralesupelec.fr}).

{\small
\bibliographystyle{ieee_fullname}
\bibliography{refs}

\begin{thebibliography}{10}\itemsep=-1pt

\bibitem{akiki2021robust}
Roland Akiki, Roger Mar{\'\i}, Carlo De~Franchis, Jean-Michel Morel, and
  Gabriele Facciolo.
\newblock Robust rational polynomial camera modelling for {SAR} and pushbroom
  imaging.
\newblock In {\em 2021 IEEE International Geoscience and Remote Sensing
  Symposium (IGARSS)}, pages 7908--7911, 2021.

\bibitem{arevalo2008shadow}
Vicente Ar{\'e}valo, Javier Gonz{\'a}lez, and Gregorio Ambrosio.
\newblock Shadow detection in colour high-resolution satellite images.
\newblock {\em International Journal of Remote Sensing}, 29(7):1945--1963,
  2008.

\bibitem{barron2021mipnerf}
Jonathan~T Barron, Ben Mildenhall, Matthew Tancik, Peter Hedman, Ricardo
  Martin-Brualla, and Pratul~P. Srinivasan.
\newblock Mip-{NeRF}: A multiscale representation for anti-aliasing neural
  radiance fields.
\newblock In {\em Proceedings of the IEEE/CVF International Conference on
  Computer Vision (ICCV)}, pages 5855--5864, 2021.

\bibitem{beyer2018ames}
Ross~A Beyer, Oleg Alexandrov, and Scott McMichael.
\newblock The {Ames Stereo Pipeline}: {NASA}'s open source software for
  deriving and processing terrain data.
\newblock {\em Earth and Space Science}, 5(9):537--548, 2018.

\bibitem{bi2020neural}
Sai Bi, Zexiang Xu, Pratul Srinivasan, Ben Mildenhall, Kalyan Sunkavalli,
  Milo{\v{s}} Ha{\v{s}}an, Yannick Hold-Geoffroy, David Kriegman, and Ravi
  Ramamoorthi.
\newblock Neural reflectance fields for appearance acquisition.
\newblock {\em arXiv preprint arXiv:2008.03824}, 2020.

\bibitem{bittner2018dsm}
Ksenia Bittner, Pablo d’Angelo, Marco K{\"o}rner, and Peter Reinartz.
\newblock {DSM-to-LoD2}: Spaceborne stereo digital surface model refinement.
\newblock {\em Remote Sensing}, 10(12):1926, 2018.

\bibitem{bittner2019late}
Ksenia Bittner, Marco K{\"o}rner, and Peter Reinartz.
\newblock Late or earlier information fusion from depth and spectral data?
  {Large}-scale digital surface model refinement by hybrid-c{GAN}.
\newblock {\em Proceedings of the IEEE/CVF Conference on Computer Vision and
  Pattern Recognition (CVPR) Workshops}, pages 1471--1478, 2019.

\bibitem{bosch2019semantic}
Marc Bosch, Kevin Foster, Gordon Christie, Sean Wang, Gregory~D Hager, and
  Myron Brown.
\newblock Semantic stereo for incidental satellite images.
\newblock In {\em 2019 IEEE Winter Conference on Applications of Computer
  Vision (WACV)}, pages 1524--1532, 2019.

\bibitem{bosch2016multiple}
Marc Bosch, Zachary Kurtz, Shea Hagstrom, and Myron Brown.
\newblock A multiple view stereo benchmark for satellite imagery.
\newblock In {\em 2016 IEEE Applied Imagery Pattern Recognition Workshop
  (AIPR)}, pages 1--9, 2016.

\bibitem{broxton20093d}
Michael~J Broxton, Ara~V Nefian, Zachary Moratto, Taemin Kim, Michael Lundy,
  and Aleksandr~V Segal.
\newblock {3D} lunar terrain reconstruction from {Apollo} images.
\newblock In {\em International Symposium on Visual Computing}, pages 710--719,
  2009.

\bibitem{dangelo2012dense}
Pablo d'Angelo and Georg Kuschk.
\newblock Dense multi-view stereo from satellite imagery.
\newblock In {\em 2012 IEEE International Geoscience and Remote Sensing
  Symposium (IGARSS)}, pages 6944--6947, 2012.

\bibitem{defranchis2014automatic}
Carlo De~Franchis, Enric Meinhardt-Llopis, Julien Michel, Jean-Michel Morel,
  and Gabriele Facciolo.
\newblock An automatic and modular stereo pipeline for pushbroom images.
\newblock {\em ISPRS Annals of the Photogrammetry, Remote Sensing and Spatial
  Information Sciences}, 2(3):49--56, 2014.

\bibitem{deng2021depth}
Kangle Deng, Andrew Liu, Jun-Yan Zhu, and Deva Ramanan.
\newblock Depth-supervised {NeRF}: Fewer views and faster training for free.
\newblock In {\em Proceedings of the IEEE/CVF Conference on Computer Vision and
  Pattern Recognition (CVPR)}, 2022.

\bibitem{derksen2021shadow}
Dawa Derksen and Dario Izzo.
\newblock Shadow neural radiance fields for multi-view satellite
  photogrammetry.
\newblock In {\em Proceedings of the IEEE/CVF Conference on Computer Vision and
  Pattern Recognition (CVPR) Workshops}, pages 1152--1161, 2021.

\bibitem{facciolo2017automatic}
Gabriele Facciolo, Carlo De~Franchis, and Enric Meinhardt-Llopis.
\newblock Automatic {3D} reconstruction from multi-date satellite images.
\newblock In {\em Proceedings of the IEEE Conference on Computer Vision and
  Pattern Recognition (CVPR) Workshops}, pages 57--66, 2017.

\bibitem{fraser2006sensor}
Clive~S Fraser, Gene Dial, and Jacek Grodecki.
\newblock Sensor orientation via {RPCs}.
\newblock {\em ISPRS Journal of Photogrammetry and Remote Sensing},
  60(3):182--194, 2006.

\bibitem{gao2021rational}
Jian Gao, Jin Liu, and Shunping Ji.
\newblock Rational polynomial camera model warping for deep learning based
  satellite multi-view stereo matching.
\newblock In {\em Proceedings of the IEEE/CVF International Conference on
  Computer Vision (ICCV)}, pages 6148--6157, 2021.

\bibitem{gomez2022experimental}
Alvaro G{\'o}mez, Gregory Randall, Gabriele Facciolo, and Rafael~Grompone von
  Gioi.
\newblock An experimental comparison of multi-view stereo approaches on
  satellite images.
\newblock In {\em Proceedings of the IEEE/CVF Winter Conference on Applications
  of Computer Vision}, pages 844--853, 2022.

\bibitem{gong2019dsm}
Ke Gong and Dieter Fritsch.
\newblock {DSM} generation from high resolution multi-view stereo satellite
  imagery.
\newblock {\em Photogrammetric Engineering \& Remote Sensing}, 85(5):379--387,
  2019.

\bibitem{grodecki2001ikonos}
Jacek Grodecki.
\newblock {IKONOS} stereo feature extraction--{RPC approach}.
\newblock In {\em ASPRS Annual Conference}, 2001.

\bibitem{grodecki2003block}
Jacek Grodecki and Gene Dial.
\newblock Block adjustment of high-resolution satellite images described by
  rational polynomials.
\newblock {\em Photogrammetric Engineering \& Remote Sensing}, 69(1):59--68,
  2003.

\bibitem{han2020state}
Yilong Han, Shugen Wang, Danchao Gong, Yue Wang, and X Ma.
\newblock State of the art in digital surface modelling from multi-view
  high-resolution satellite images.
\newblock {\em ISPRS Annals of the Photogrammetry, Remote Sensing and Spatial
  Information Sciences}, 5(2):351--356, 2020.

\bibitem{hedman2021baking}
Peter Hedman, Pratul~P Srinivasan, Ben Mildenhall, Jonathan~T Barron, and Paul
  Debevec.
\newblock Baking neural radiance fields for real-time view synthesis.
\newblock In {\em Proceedings of the IEEE/CVF International Conference on
  Computer Vision (ICCV)}, pages 5875--5884, 2021.

\bibitem{hirschmuller2008stereo}
Heiko Hirschmuller.
\newblock Stereo processing by semiglobal matching and mutual information.
\newblock {\em IEEE Transactions on Pattern Analysis and Machine Intelligence},
  30(2):328--341, 2008.

\bibitem{jain2021putting}
Ajay Jain, Matthew Tancik, and Pieter Abbeel.
\newblock Putting {NeRF} on a diet: Semantically consistent few-shot view
  synthesis.
\newblock In {\em Proceedings of the IEEE/CVF International Conference on
  Computer Vision (ICCV)}, pages 5885--5894, 2021.

\bibitem{kendall2018multi}
Alex Kendall, Yarin Gal, and Roberto Cipolla.
\newblock Multi-task learning using uncertainty to weigh losses for scene
  geometry and semantics.
\newblock In {\em Proceedings of the IEEE Conference on Computer Vision and
  Pattern Recognition (CVPR)}, pages 7482--7491, 2018.

\bibitem{le20192019}
Bertrand Le~Saux, Naoto Yokoya, Ronny Hansch, Myron Brown, and Greg Hager.
\newblock 2019 data fusion contest [technical committees].
\newblock {\em IEEE Geoscience and Remote Sensing Magazine}, 7(1):103--105,
  2019.

\bibitem{liebel2020generalized}
Lukas Liebel, Ksenia Bittner, and Marco K{\"o}rner.
\newblock A generalized multi-task learning approach to stereo {DSM} filtering
  in urban areas.
\newblock {\em ISPRS Journal of Photogrammetry and Remote Sensing},
  166:213--227, 2020.

\bibitem{lowe2004distinctive}
David~G Lowe.
\newblock Distinctive image features from scale-invariant keypoints.
\newblock {\em International Journal of Computer Vision}, 60(2):91--110, 2004.

\bibitem{ma2008shadow}
Haijian Ma, Qiming Qin, and Xinyi Shen.
\newblock Shadow segmentation and compensation in high resolution satellite
  images.
\newblock In {\em 2008 IEEE International Geoscience and Remote Sensing
  Symposium (IGARSS)}, volume~2, pages 1036--1039, 2008.

\bibitem{mari2021generic}
Roger Mar{\'\i}, Carlo {de Franchis}, Enric Meinhardt-Llopis, J{\'e}r{\'e}my
  Anger, and Gabriele Facciolo.
\newblock {A generic bundle adjustment methodology for indirect RPC model
  refinement of satellite imagery}.
\newblock {\em Image Processing On Line}, 11:344--373, 2021.

\bibitem{mari2019bundle}
Roger Mar{\'\i}, Carlo de Franchis, Enric Meinhardt-Llopis, and Gabriele
  Facciolo.
\newblock To bundle adjust or not: A comparison of relative geolocation
  correction strategies for satellite multi-view stereo.
\newblock In {\em Proceedings of the IEEE/CVF International Conference on
  Computer Vision (ICCV) Workshops}, 2019.

\bibitem{martin2021nerf}
Ricardo Martin-Brualla, Noha Radwan, Mehdi~SM Sajjadi, Jonathan~T Barron,
  Alexey Dosovitskiy, and Daniel Duckworth.
\newblock {NeRF} in the wild: Neural radiance fields for unconstrained photo
  collections.
\newblock In {\em Proceedings of the IEEE/CVF Conference on Computer Vision and
  Pattern Recognition (CVPR)}, pages 7210--7219, 2021.

\bibitem{mildenhall2021nerf}
Ben Mildenhall, Peter Hedman, Ricardo Martin-Brualla, Pratul Srinivasan, and
  Jonathan~T Barron.
\newblock {NeRF} in the dark: High dynamic range view synthesis from noisy raw
  images.
\newblock In {\em Proceedings of the IEEE/CVF Conference on Computer Vision and
  Pattern Recognition (CVPR)}, 2022.

\bibitem{mildenhall2020nerf}
Ben Mildenhall, Pratul~P Srinivasan, Matthew Tancik, Jonathan~T Barron, Ravi
  Ramamoorthi, and Ren Ng.
\newblock {NeRF}: Representing scenes as neural radiance fields for view
  synthesis.
\newblock In {\em European Conference on Computer Vision}, pages 405--421,
  2020.

\bibitem{muller2022instant}
Thomas M{\"u}ller, Alex Evans, Christoph Schied, and Alexander Keller.
\newblock Instant neural graphics primitives with a multiresolution hash
  encoding.
\newblock {\em arXiv preprint arXiv:2201.05989}, 2022.

\bibitem{ozcanli2015comparison}
Ozge~C Ozcanli, Yi Dong, Joseph~L Mundy, Helen Webb, Riad Hammoud, and Victor
  Tom.
\newblock A comparison of stereo and multiview {3-D} reconstruction using
  cross-sensor satellite imagery.
\newblock In {\em Proceedings of the IEEE Conference on Computer Vision and
  Pattern Recognition (CVPR) Workshops}, pages 17--25, 2015.

\bibitem{ozcanli2014automatic}
Ozge~C Ozcanli, Yi Dong, Joseph~L Mundy, Helen Webb, Riad Hammoud, and Tom
  Victor.
\newblock Automatic geo-location correction of satellite imagery.
\newblock In {\em Proceedings of the IEEE Conference on Computer Vision and
  Pattern Recognition (CVPR) Workshops}, pages 307--314, 2014.

\bibitem{pan2021self}
Hongbo Pan, Tao Huang, Ping Zhou, and Zehua Cui.
\newblock Self-calibration dense bundle adjustment of multi-view {Worldview-3}
  basic images.
\newblock {\em ISPRS Journal of Photogrammetry and Remote Sensing},
  176:127--138, 2021.

\bibitem{park2021nerfies}
Keunhong Park, Utkarsh Sinha, Jonathan~T Barron, Sofien Bouaziz, Dan~B Goldman,
  Steven~M Seitz, and Ricardo Martin-Brualla.
\newblock Nerfies: Deformable neural radiance fields.
\newblock In {\em Proceedings of the IEEE/CVF International Conference on
  Computer Vision (ICCV)}, pages 5865--5874, 2021.

\bibitem{park2021hypernerf}
Keunhong Park, Utkarsh Sinha, Peter Hedman, Jonathan~T. Barron, Sofien Bouaziz,
  Dan~B Goldman, Ricardo Martin-Brualla, and Steven~M. Seitz.
\newblock {HyperNeRF}: A higher-dimensional representation for topologically
  varying neural radiance fields.
\newblock {\em ACM Trans. Graph.}, 40(6), 2021.

\bibitem{pumarola2021d}
Albert Pumarola, Enric Corona, Gerard Pons-Moll, and Francesc Moreno-Noguer.
\newblock {D-NeRF}: Neural radiance fields for dynamic scenes.
\newblock In {\em Proceedings of the IEEE/CVF Conference on Computer Vision and
  Pattern Recognition (CVPR)}, pages 10318--10327, 2021.

\bibitem{roessle2021dense}
Barbara Roessle, Jonathan~T Barron, Ben Mildenhall, Pratul~P Srinivasan, and
  Matthias Nie{\ss}ner.
\newblock Dense depth priors for neural radiance fields from sparse input
  views.
\newblock In {\em Proceedings of the IEEE/CVF Conference on Computer Vision and
  Pattern Recognition (CVPR)}, 2022.

\bibitem{rupnik20183d}
Ewelina Rupnik, Marc Pierrot-Deseilligny, and Arthur Delorme.
\newblock {3D reconstruction from multi-view VHR-satellite images in MicMac}.
\newblock {\em ISPRS Journal of Photogrammetry and Remote Sensing},
  139:201--211, 2018.

\bibitem{shean2016automated}
David~E Shean, Oleg Alexandrov, Zachary~M Moratto, Benjamin~E Smith, Ian~R
  Joughin, Claire Porter, and Paul Morin.
\newblock An automated, open-source pipeline for mass production of digital
  elevation models ({DEMs}) from very-high-resolution commercial stereo
  satellite imagery.
\newblock {\em ISPRS Journal of Photogrammetry and Remote Sensing},
  116:101--117, 2016.

\bibitem{sitzmann2020implicit}
Vincent Sitzmann, Julien Martel, Alexander Bergman, David Lindell, and Gordon
  Wetzstein.
\newblock Implicit neural representations with periodic activation functions.
\newblock {\em Advances in Neural Information Processing Systems},
  33:7462--7473, 2020.

\bibitem{srinivasan2021nerv}
Pratul~P Srinivasan, Boyang Deng, Xiuming Zhang, Matthew Tancik, Ben
  Mildenhall, and Jonathan~T Barron.
\newblock {NeRV}: Neural reflectance and visibility fields for relighting and
  view synthesis.
\newblock In {\em Proceedings of the IEEE/CVF Conference on Computer Vision and
  Pattern Recognition (CVPR)}, pages 7495--7504, 2021.

\bibitem{stucker2020resdepth}
Corinne Stucker and Konrad Schindler.
\newblock {ResDepth}: Learned residual stereo reconstruction.
\newblock In {\em Proceedings of the IEEE/CVF Conference on Computer Vision and
  Pattern Recognition (CVPR) Workshops}, pages 184--185, 2020.

\bibitem{tewari2020state}
Ayush Tewari, Ohad Fried, Justus Thies, Vincent Sitzmann, Stephen Lombardi,
  Kalyan Sunkavalli, Ricardo Martin-Brualla, Tomas Simon, Jason Saragih,
  Matthias Nie{\ss}ner, et~al.
\newblock State of the art on neural rendering.
\newblock In {\em Computer Graphics Forum}, volume~39, pages 701--727, 2020.

\bibitem{triggs1999bundle}
Bill Triggs, Philip~F McLauchlan, Richard~I Hartley, and Andrew~W Fitzgibbon.
\newblock Bundle adjustment—a modern synthesis.
\newblock In {\em International Workshop on Vision Algorithms}, pages 298--372,
  1999.

\bibitem{xu2021h}
Hongyi Xu, Thiemo Alldieck, and Cristian Sminchisescu.
\newblock {H-NeRF}: Neural radiance fields for rendering and temporal
  reconstruction of humans in motion.
\newblock {\em Advances in Neural Information Processing Systems}, 34, 2021.

\bibitem{yang2021recursive}
Guo-Wei Yang, Wen-Yang Zhou, Hao-Yang Peng, Dun Liang, Tai-Jiang Mu, and
  Shi-Min Hu.
\newblock Recursive-{NeRF}: An efficient and dynamically growing {NeRF}.
\newblock {\em arXiv preprint arXiv:2105.09103}, 2021.

\bibitem{yu2021plenoctrees}
Alex Yu, Ruilong Li, Matthew Tancik, Hao Li, Ren Ng, and Angjoo Kanazawa.
\newblock {PlenOctrees} for real-time rendering of neural radiance fields.
\newblock In {\em Proceedings of the IEEE/CVF International Conference on
  Computer Vision (ICCV)}, pages 5752--5761, 2021.

\bibitem{yu2021pixelnerf}
Alex Yu, Vickie Ye, Matthew Tancik, and Angjoo Kanazawa.
\newblock {pixelNeRF}: Neural radiance fields from one or few images.
\newblock In {\em Proceedings of the IEEE/CVF Conference on Computer Vision and
  Pattern Recognition (CVPR)}, pages 4578--4587, 2021.

\bibitem{zhang2019leveraging}
Kai Zhang, Noah Snavely, and Jin Sun.
\newblock Leveraging vision reconstruction pipelines for satellite imagery.
\newblock In {\em Proceedings of the IEEE/CVF International Conference on
  Computer Vision (ICCV) Workshops}, 2019.

\end{thebibliography}
}

\end{document}